\documentclass{article}

\usepackage{arxiv}

\usepackage[utf8]{inputenc}
\usepackage[T1]{fontenc}
\usepackage{hyperref}
\usepackage{url}
\usepackage{booktabs}
\usepackage{amsfonts}
\usepackage{nicefrac}
\usepackage{microtype}
\usepackage{graphicx}
\usepackage{natbib}
\usepackage{doi}

\usepackage{subcaption}
\usepackage{array}

\title{Astro-HEP-BERT: \\ A bidirectional language model for studying the meanings of concepts in astrophysics and high energy physics}

\author{\href{https://orcid.org/0000-0003-0657-5254}{\includegraphics[scale=0.06]{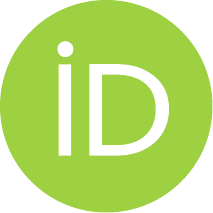}\hspace{1mm}Arno Simons} \\ History and Philosophy of Modern Science \\ Technische Universität Berlin \\ Straße des 17. Juni 135, 10623 Berlin \\ \texttt{arno.simons@gmail.com}}

\renewcommand{\headeright}{}
\renewcommand{\undertitle}{}
\renewcommand{\shorttitle}{Astro-HEP-BERT}

\hypersetup{
    pdftitle={Astro-HEP-BERT: A bidirectional language model for studying the meanings of concepts in astrophysics and high energy physics},
    pdfsubject={cs.CL, physics.hist-ph},
    pdfauthor={Arno Simons},
    pdfkeywords={scientific concepts, contextualized word embeddings, history of science, philosophy of science, sociology of science, word sense disambiguation and induction, lexical semantic change, discourse analysis},
}

\begin{document}
\maketitle

\begin{abstract}
I present \texttt{Astro-HEP-BERT}, a transformer-based language model specifically designed for generating contextualized word embeddings (CWEs) to study the meanings of concepts in astrophysics and high-energy physics. Built on a general pretrained \texttt{BERT} model, \texttt{Astro-HEP-BERT} underwent further training over three epochs using the Astro-HEP Corpus, a dataset I curated from 21.84 million paragraphs extracted from more than 600,000 scholarly articles on arXiv, all belonging to at least one of these two scientific domains. The project demonstrates both the effectiveness and feasibility of adapting a bidirectional transformer for applications in the history, philosophy, and sociology of science (HPSS). The entire training process was conducted using freely available code, pretrained weights, and text inputs, completed on a single MacBook Pro Laptop (M2/96GB). Preliminary evaluations indicate that \texttt{Astro-HEP-BERT}'s CWEs perform comparably to domain-adapted BERT models trained from scratch on larger datasets for domain-specific word sense disambiguation and induction and related semantic change analyses. This suggests that retraining general language models for specific scientific domains can be a cost-effective and efficient strategy for HPSS researchers, enabling high performance without the need for extensive training from scratch.
\end{abstract}

\keywords{
    scientific concepts \and
    contextualized word embeddings \and
    history, philosophy, and sociology of science \and
    word sense disambiguation and induction \and
    lexical semantic change \and
    discourse analysis
    }

\section{Introduction}
\label{sec:introduction}

Understanding \textbf{the meanings of scientific concepts}, their evolution over time, and their differing interpretations across research fields is a central focus in the \textbf{history, philosophy, and sociology of science (HPSS)}. HPSS researchers have addressed these questions using both qualitative and quantitative methodologies. Traditional \textbf{``close reading''} approaches in HPSS provide detailed, context-rich analyses of conceptual histories, often yielding profound philosophical insights \citep{steinle_exploratory_2016, chang_inventing_2007, pickering_constructing_1999, fleck_genesis_1979, hacking_emergence_1975}. However, these studies tend to be time-intensive, difficult to replicate, and constrained in scope, which can limit their broader applicability and generalizability.

More recent \textbf{``distant reading''} approaches have introduced co-occurrence-based methods---including latent semantic analysis, latent Dirichlet allocation, and word2vec---to investigate the meanings and historical trajectories of concepts within large scientific corpora. These methods reveal connections and patterns across texts that close reading alone might miss, enabling broader analyses across expansive datasets \citep{malaterre_epistemic_2024, lean_digital_2023, wevers_digital_2020, laubichler_computational_2019, pence_how_2018, overton_explain_2013, callon_qualitative_1986}. However, a limitation of these approaches is their often simplified representation of meaning, typically overlooking the nuances introduced by word order and cases of colexification, where one word can carry multiple meanings. This trade-off often results in an incomplete analysis that sacrifices depth for breadth.

\textbf{Contextualized word embeddings (CWEs)}, especially in models trained bidirectionally, such as BERT (Bidirectional Encoder Representations from Transformers) \citep{devlin_bert:_2018}, offer a promising synthesis of the close and distant reading methods while addressing their respective limitations \citep{simons_meaning_2024, zichert_tracing_2024, kleymann_conceptual_2022}. Unlike static word embeddings \citep{mikolov_efficient_2013}, CWEs generate unique vector representations for each word based on its specific context of occurrence. This enables CWEs to capture meaning on both a global scale, similar to static embeddings that reflect the domain-specific training corpus, and on a local scale, where word meanings are adapted to their immediate textual surroundings.

Due to their sensitivity to local context, \textbf{CWEs can handle colexification}, providing an effective approach to interpreting words with multiple meanings. Examples include homographs like ``light'', which could refer to weight in the particle is ``\textit{light}'' or to the phenomenon in ``\textit{light} is a particle''. Additionally, CWEs manage polysemous words—terms with dominant and subordinate meanings---such as ``\textit{deep},'' which generally implies extending far down but could also refer to ``Hubble \textit{Deep} Field''' or ``\textit{deep} learning''. Consequently, CWEs can be leveraged to track concept evolution, explore semantic relationships, and differentiate between dominant and subordinate meanings \citep{periti_lexical_2024, loureiro_language_2020, wiedemann_does_2019}.

Research has consistently shown that CWEs are significantly improved when pretrained on \textbf{domain-specific corpora}. Prominent examples include \texttt{BioBERT} \citep{lee_biobert_2020}, \texttt{SciBERT} \citep{beltagy_scibert_2019}, and \texttt{PubMedBERT} \citep{gu_domain-specific_2021} in the biomedical field, which set new benchmarks in biomedical NLP. In physics, notable models such as \texttt{PhysBERT} \citep{hellert_physbert_2024} and \texttt{astroBERT} \citep{grezes_improving_2022, grezes_building_2021} were trained from scratch on large physics corpora, yielding models tuned for physics-specific text.

This paper introduces both the \textbf{Astro-HEP-BERT} model and the \textbf{Astro-HEP Corpus}. \texttt{Astro-HEP-BERT} adapts a general-purpose BERT model to astrophysics and HEP through three additional training epochs on the Astro-HEP Corpus, which includes over 21.84 million paragraphs from more than 600,000 scholarly articles in these fields. Unlike \texttt{PhysBERT} and \texttt{astroBERT}, which were trained from scratch, \texttt{Astro-HEP-BERT} leverages pre-existing linguistic patterns and vocabulary from the general-purpose BERT model, enabling targeted domain adaptation with reduced computational demands.

The \texttt{Astro-HEP-BERT} model is publicly \textbf{available on Hugging Face} under the handle \href{https://huggingface.co/arnosimons/astro-hep-bert}{``arnosimons/astro-hep-bert''}\footnote{\url{https://huggingface.co/arnosimons/astro-hep-bert}}, along with sample code for extracting word embeddings. Additionally, the Astro-HEP Corpus is described on Hugging Face in the form of a dataset card under the name \href{https://huggingface.co/datasets/arnosimons/astro-hep-corpus}{``arnosimons/astro-hep-corpus''}\footnote{\url{https://huggingface.co/datasets/arnosimons/astro-hep-corpus}}.

In \textbf{a complementary paper} \citep{simons_meaning_2024} I evaluate the performance of \texttt{Astro-HEP-BERT} in comparison to four other BERT-based models---\texttt{PhysBERT}, \texttt{astroBERT}, \texttt{SciBERT}, and \texttt{BERT}---in disambiguating and tracking the meaning of the term ``Planck'' over a 30 year period.

\section{The Astro-HEP Corpus}
\label{sec:corpus}

\begin{figure}[htp!]
    \centering
    \includegraphics[width=13cm]{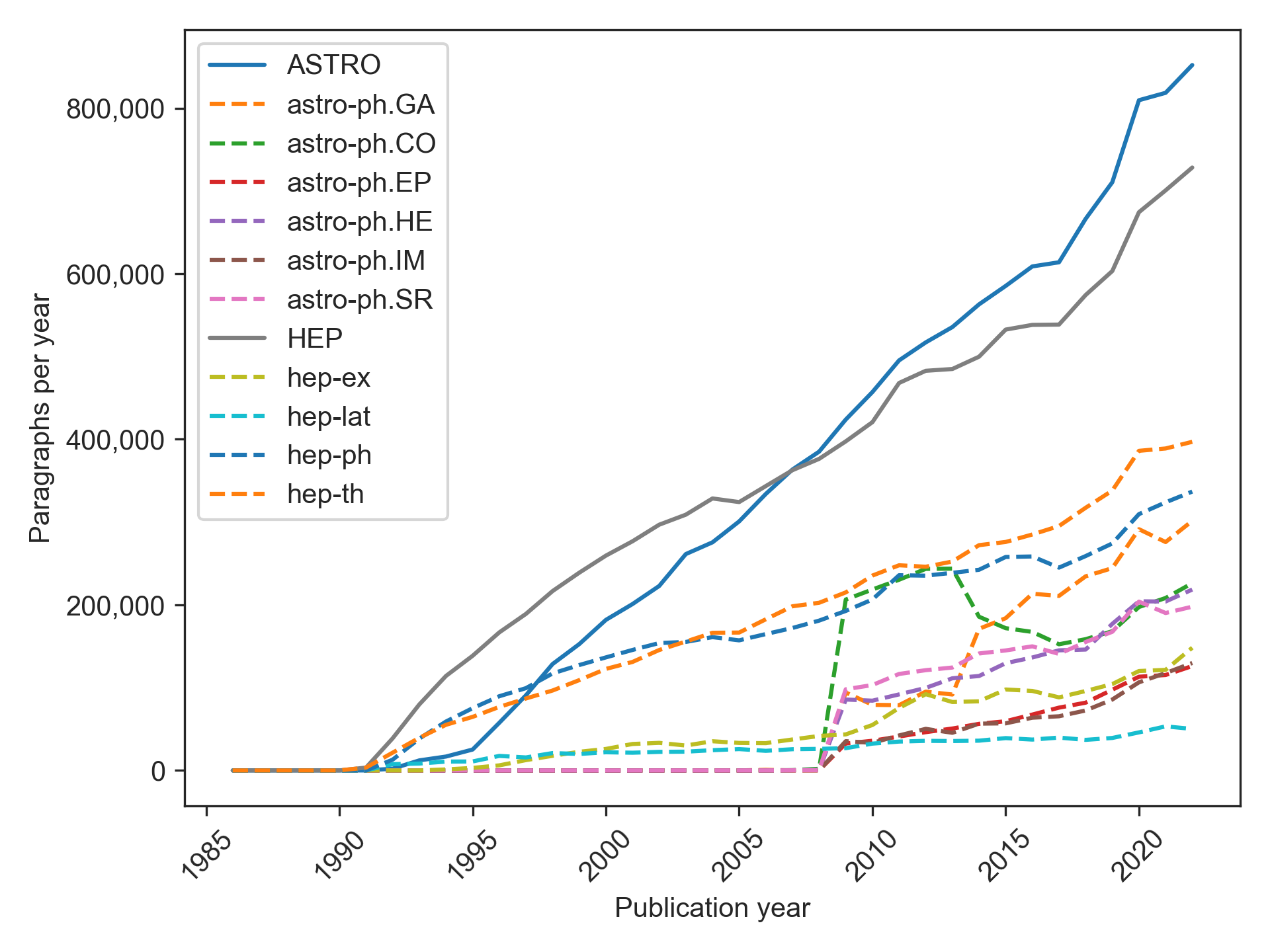}
    \caption{The Astro-HEP Corpus: 21.84M paragraphs found in 0.61M articles on astrophysics (ASTRO) and/or high energy physics (HEP) published between 1986 and 2022 on arXiv.}
    \label{fig:corpus}
\end{figure}

The Astro-HEP Corpus consists of approximately \textbf{21.8 million paragraphs} extracted from over 600,000 scholarly articles on astrophysics and HEP, all \textbf{published between 1986 and 2022} on the open-access archive \href{http://arxiv.org/}{arXiv.org}. Figure \ref{fig:corpus} illustrates the temporal distribution of articles based on their publication year, category (ASTRO or HEP), and subcategory, with metadata directly imported \textbf{from arXiv.org}.

To compile the corpus, I obtained the original arXiv metadata file from the \href{https://www.kaggle.com/datasets/Cornell-University/arxiv}{Kaggle platform}\footnote{\url{https://www.kaggle.com/datasets/Cornell-University/arxiv}} and filtered it using the official \href{https://arxiv.org/category_taxonomy}{arXiv taxonomy}\footnote{\url{https://arxiv.org/category_taxonomy}}. This taxonomy categorizes high-energy physics into four areas (``hep-ex'', ``hep-lat'', ``hep-ph'', and ``hep-th'') and astrophysics into six subcategories (``astro-ph.CO'', ``astro-ph.EP'', ``astro-ph.GA'', ``astro-ph.HE'', ``astro-ph.IM'', and ``astro-ph.SR''). Using arXiv IDs, I then downloaded the corresponding zipped \textbf{source files}, which mainly contained \textbf{LaTeX documents}, and retained only these for further processing.

To convert these LaTeX documents into \textbf{plain text}, I used \href{https://pandoc.org/}{Pandoc}\footnote{\url{https://pandoc.org/}}, an open-source document converter. To optimize parsing accuracy, I integrated Pandoc into a Python script that pre-processed the LaTeX code, standardizing atypical commands, spacing, and symbols before conversion. Afterward, the script refined the text further, \textbf{replacing in-text citations} with a ``[CIT]'' marker \textbf{and multi-line mathematical expressions} with a ``FORMULA'' marker, while inline mathematical expressions (``$\dots$'') were retained. The final step involved parsing the plain text into individual paragraphs using newline splitting.

\begin{figure}[htp!]
    \centering
    \includegraphics[width=10cm]{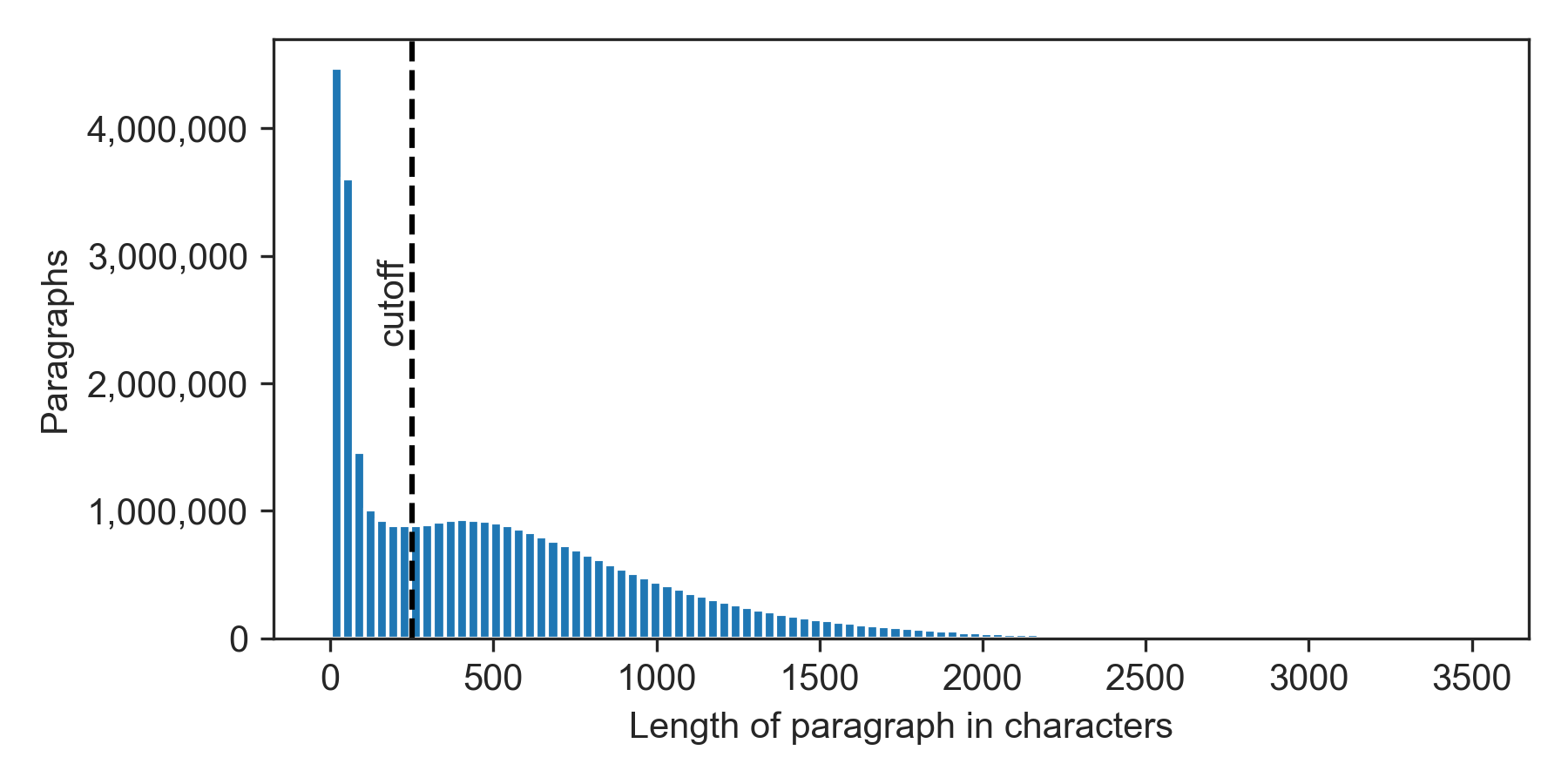}
    \caption{Distribution of paragraph length before filtering out short paragraphs---35.38M paragraphs found in 0.61M articles on astrophysics (ASTRO) and/or high energy physics (HEP) published between 1986 and 2022 on arXiv.}
    \label{fig:article-length}
\end{figure}

\begin{figure}[htp!]
    \centering
    \includegraphics[width=10cm]{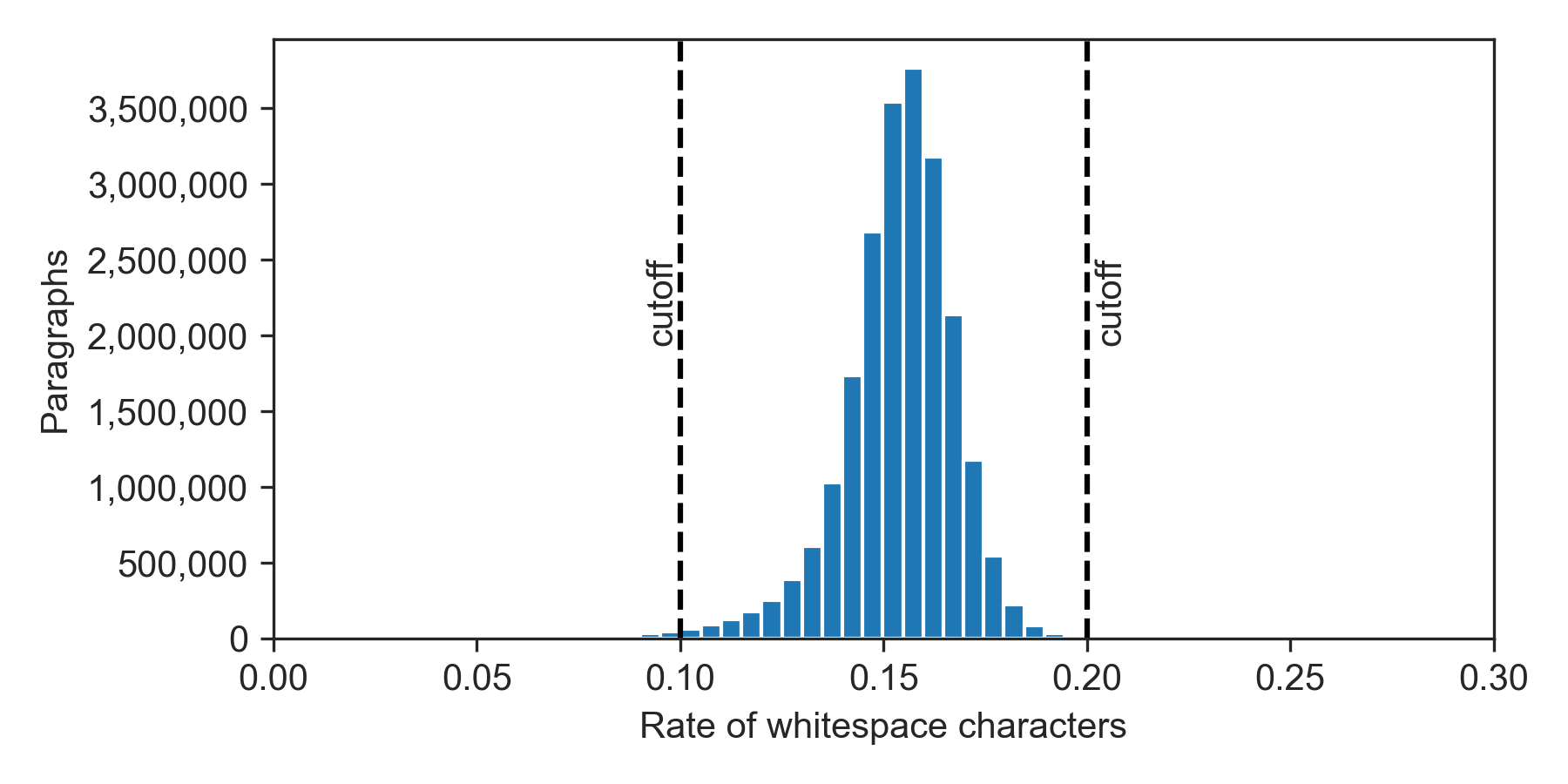}
    \caption{Distribution of whitespace rate before filtering out paragraphs with a rate of less than 0.1 or more than 0.2---22.03M paragraphs found in 0.61M articles on astrophysics (ASTRO) and/or high energy physics (HEP) published between 1986 and 2022 on arXiv.}
    \label{fig:article-whitespace}
\end{figure}

Despite these steps, some non-paragraph artifacts, like contact details or citations, remained due to uncommon LaTeX formatting. Occasionally, gibberish lines also appeared from the parsing process. I used two strategies to filter out these artifacts:

\begin{quote}
    \textbf{Filtering Short Paragraphs}: After conducting a frequency analysis of paragraph length (shown in Figure \ref{fig:article-length}), I defined a lower cutoff of 250 characters. This step reduced the number of paragraphs from 35.38 million to 22.03 million.
\end{quote}
\begin{quote}
    \textbf{Whitespace Character Rate Filtering}: Further filtering removed paragraphs with atypical whitespace rates, identified through a frequency analysis combined with manual sample inspection (see Figure \ref{fig:article-whitespace}). I set thresholds at 0.1 for low whitespace rates and 0.2 for high rates, reducing the corpus from 22.03 million to 21.84 million paragraphs.
\end{quote}

Finally, I merged the filtered paragraphs with relevant metadata into a unified dataframe, which includes the columns described in Table \ref{table:columns}.

\begin{table}[h!]
\centering
\begin{tabular}{  l l  }
 \toprule
 Column & Description  \\ 
 \midrule
 \textit{Text} & Full text of the paragraph\\
 \textit{Characters} & Number of unicode characters in the paragraph\\
 \textit{Subwords} & Number of subwords in the paragraph\\
 \textit{arXiv ID} & Identifier of the parent article provided by arXiv\\
 \textit{Year} & Year of the first publication of the parent article\\
 \textit{Month} & Month of the first publication of the parent article\\
 \textit{Day} & Day of the first publication of the parent article\\
 \textit{Position} & Position in the sequence of paragraphs in the article \\
 \bottomrule
\end{tabular}
\caption{Main columns of the Astro-HEP Corpus}
\label{table:columns}
\end{table}

\section{The Astro-HEP-BERT Model}
\label{sec:model}

\texttt{Astro-HEP-BERT} is a bidirectional transformer designed to generate contextualized word embeddings for analyzing the meanings of concepts in astrophysics and high-energy physics. Built on the uncased base version of \texttt{BERT} \citep{devlin_bert:_2018}, it \textbf{reuses BERT's pretrained weights and vocabulary}, followed by an additional three epochs of domain-specific pre-training on the Astro-HEP Corpus.

To optimize \texttt{Astro-HEP-BERT} for this specialized corpus, I introduced several enhancements to the original BERT training protocol. First, in line with findings by \citet{mickus_what_2020} and \citet{liu_roberta_2019}, I removed the Next Sentence Prediction (NSP) objective, focusing solely on \textbf{Masked Language Modeling (MLM)} as the training objective. Second, I applied \textbf{whole-word masking} \citep{devlin_bert_2019}, ensuring that when a word consists of multiple subwords, all subwords are masked consistently to maintain a uniform masking rate. Third, to maximize semantic consistency within training examples, I refined the document-sentence input format proposed by \citet{liu_roberta_2019} by restricting sequences to complete, individual paragraphs--a structure I term the \textbf{full-paragraphs format}. This format recognizes the paragraph as the basic unit of meaning in academic writing, where a single, cohesive idea is typically developed across several sentences. Since training BERT with document sentences already shows improvements over single-sentence training; by using full-paragraphs, I anticipate even stronger semantic coherence in the model's contextualized embeddings.

To account for the wide variation in paragraph lengths, ranging from 48 to 510 subwords, I organized \textbf{training batches} to achieve \textbf{a uniform token count of 8,192}. Rather than using a fixed example count per batch, this approach ensured that 1) the number of padding tokens in each batch did not exceed 20\% of total tokens, and 2) each batch's token count fell within 8,192$\pm410$ (5\% of the target). By minimizing padding—tokens that act as placeholders without meaningful information---this procedure allocates more computational effort to processing actual data, enhancing overall efficiency.

Over \textbf{three training epochs}, \texttt{Astro-HEP-BERT} processed a total of 12.7 billion tokens, of which 1.9 billion were masked. Training took 48 days on a single MacBook Pro equipped with an M2 chip and 96GB of memory. These results demonstrate the feasibility of training a custom bidirectional transformer on a large, domain-specific corpus as an open-source project with relatively modest computational resources.

\begin{figure}[htp!]
    \centering
    \includegraphics[width=12
cm]{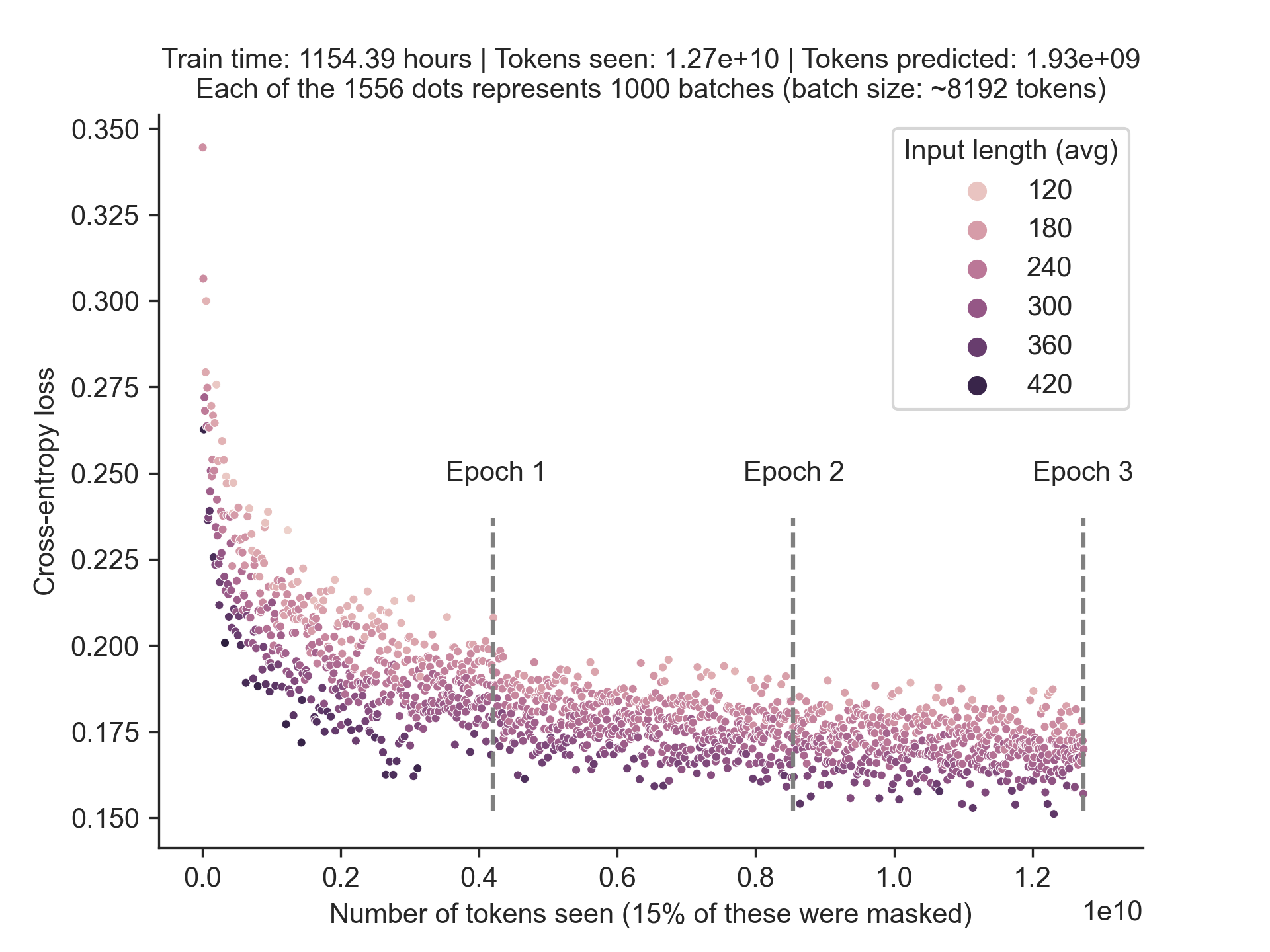}
    \caption{Decreasing cross-entropy loss during the extended pretraining of \texttt{Astro-HEP-BERT}.}
    \label{fig:loss}
\end{figure}

As shown in Figure \ref{fig:loss}, \texttt{Astro-HEP-BERT} displayed a continuous reduction in training loss over time, indicating improved prediction accuracy for masked words. The decreasing cross-entropy loss demonstrates that \texttt{Astro-HEP-BERT} developed an increasingly robust understanding of the specialized language within the training data compared to the original \texttt{BERT} model.

\section{Evaluation}
\label{sec:evaluation}

In \citet{simons_meaning_2024}, I evaluate the performance of \textbf{five BERT-based models}, including my own \texttt{Astro-HEP-BERT}, each varying in the degree of domain-specific pretraining, on tasks such as \textbf{word sense disambiguation, induction}, and \textbf{semantic change detection}, using the polysemous physics term \textbf{``Planck'' as a test study}. My results show that the domain-adapted models---\texttt{Astro-HEP-BERT}, \texttt{PhysBERT}, and \texttt{astroBERT}---better predict the known meanings of ``Planck'', produce cleaner word sense clusters, and are thereby able to identify key shifts in the term's meaning coinciding with the emergence and growing relevance of the Planck space mission. 

\section{Conclusion}

This paper introduced \textbf{Astro-HEP-BERT}, a bidirectional transformer model tailored for understanding astrophysics and high-energy physics (HEP) language. Building on the general-purpose \texttt{BERT} base model and fine-tuned with the newly developed Astro-HEP Corpus, \texttt{Astro-HEP-BERT} combines the computational efficiency of pretrained models with specialized understanding for scientific discourse. 

The study illustrates that developing \textbf{customized bidirectional transformers for specialized scientific languages} is both feasible and accessible. Using only freely available code, pretrained weights, and open-source data, training was completed on a single MacBook Pro (M2/96GB), showing that effective domain adaptation does not require extensive resources. 

Domain-specific transformers like \texttt{Astro-HEP-BERT} open \textbf{new possibilities for} studying meanings of scientific concepts within the history, philosophy, and sociology of science (\textbf{HPSS}), capturing nuanced meanings in context with vector representations that reflect semantic relationships. My evaluation results in \citet{simons_meaning_2024} demonstrate that in disambiguating and tracking the different meanings of the term ``Planck'' \texttt{Astro-HEP-BERT} performs comparably with four leading BERT models, two of which trained from scratch on extensive physics corpora.

\textbf{Keeping pace with advancements} in areas such as ``BERTology'' \citep{rogers_primer_2020}, word sense disambiguation and induction \citep{sun_method_2023, loureiro_language_2020}, semantic change detection \citep{periti_lexical_2024}, and ``digital Begriffsgeschichte'' \citep{wevers_digital_2020} will be essential to unlock the full potential of \texttt{Astro-HEP-BERT} and similar models for analyzing scientific discourse.

\section*{Acknowledgements}

I am grateful to my colleagues \textbf{Adrian Wüthrich} and \textbf{Michael Zichert} for their insightful feedback on this draft, and for our collaborative exchange of ideas regarding the use of computational methods in HPSS. I also want to acknowledge funding by \textbf{the European Union} (ERC Consolidator Grant, Project No. 101044932). Views and opinions expressed are however those of the author only and do not necessarily reflect those of the European Union or the European Research Council. Neither the European Union nor the granting authority can be held responsible for them.

\bibliographystyle{apalike}
\bibliography{main}  

\end{document}